\pdfoutput=1

\documentclass[11pt]{article}

\usepackage[preprint]{coling}

\usepackage{times}
\usepackage{latexsym}
\usepackage{times,latexsym}
\usepackage{url}
\usepackage{hyperref}
\usepackage{multirow}
\usepackage{expex}
\usepackage[colorinlistoftodos]{todonotes}
\usepackage{amsmath}
\usepackage{graphicx}

\usepackage[normalem]{ulem}
\useunder{\uline}{\ul}{}

\usepackage{color-edits}

\usepackage[T1]{fontenc}

\usepackage[utf8]{inputenc}
\usepackage{microtype}
\usepackage{inconsolata}
\usepackage{graphicx}

\title{Boosting the Capabilities of Compact Models 
in Low-Data Contexts 
\\
with Large Language Models and Retrieval-Augmented Generation}

\author{Bhargav Shandilya and
  Alexis Palmer \\
  University of Colorado Boulder\\
\texttt{\{bhargav.shandilya,alexis.palmer\}@colorado.edu}
}

\begin{document}
\maketitle
\begin{abstract}
The data and compute requirements of current language modeling technology pose challenges for the processing and analysis of low-resource languages. Declarative linguistic knowledge has the potential to partially bridge this data scarcity gap by providing models with useful inductive bias in the form of language-specific rules. 
In this paper, we propose a retrieval augmented generation (RAG) framework backed by a large language model (LLM) to correct the output of a smaller model for the linguistic task of morphological glossing.
We leverage linguistic information to make up for the lack of data and trainable parameters, while allowing for inputs from written descriptive grammars interpreted and distilled through an LLM. 

The results demonstrate that significant leaps in performance and efficiency are possible with the right combination of: a) linguistic inputs in the form of grammars, b) the interpretive power of LLMs, and c) the trainability of smaller token classification networks. We show that a compact, RAG-supported model is highly effective in data-scarce settings, achieving a new state-of-the-art for this task and our target languages. 
Our work also offers documentary linguists a more reliable and more usable tool for morphological glossing by providing well-reasoned explanations and confidence scores for each output. \footnote{Code and data samples are available and will be released upon publication.}
\end{abstract}

\section{Introduction}

\begin{figure}[t]
  \centering
   \includegraphics[width=\linewidth]{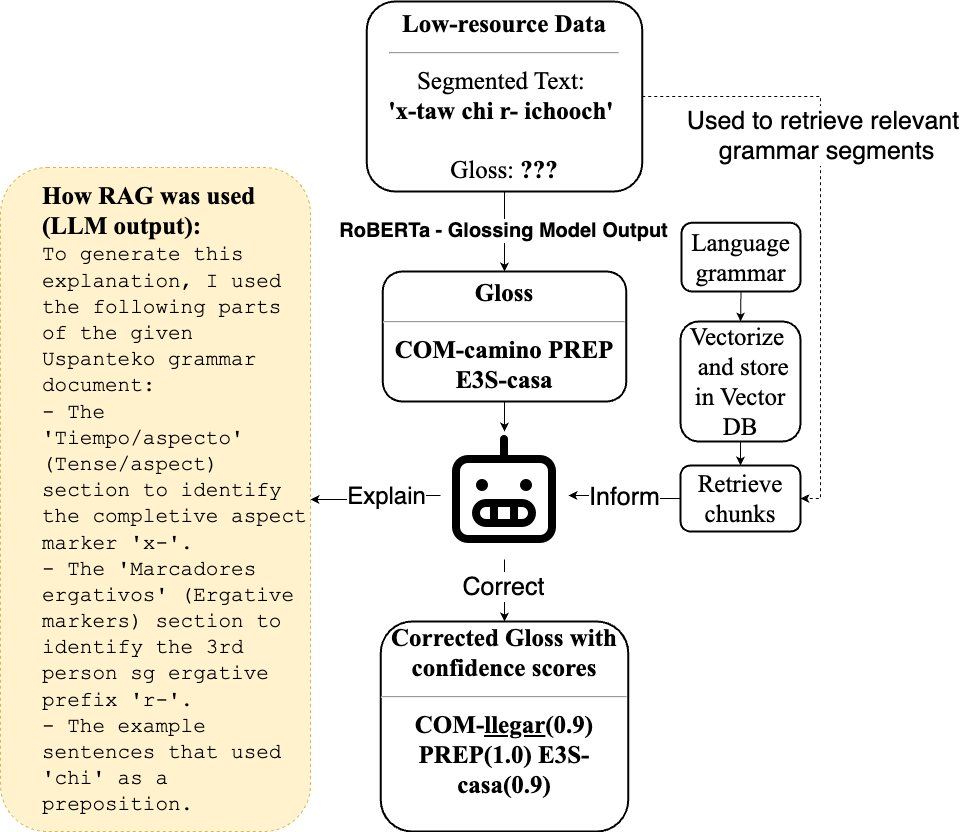}
  \caption{One Uspanteko sentence, with its original gloss, a predicted gloss, and an explanation from our IGT-RAG model. For each morpheme in the sentence, the model describes which section of the provided Uspanteko grammar it used to make its labeling decision.} 
  \label{fig:explain-ex}
\end{figure}

Over the last decade, language models have evolved rapidly, culminating in impressively domain-agnostic decoder-only models like GPT \cite{DBLP:journals/corr/abs-2005-14165} and Llama 2 \cite{touvron2023llama}. Although these models can be versatile in terms of being applicable to a wide variety of tasks and providing straightforward interfaces for quick inference, the fact remains that they are extremely parameter-heavy, making them difficult and expensive to train \cite{bender-koller-2020-climbing}. LLMs, however, give us a unique descriptive power that can boost explainability when used in certain contexts. In this paper, we examine how LLMs can be used to make RAG-informed corrections for the task of morpheme glossing (Section~\ref{sec:igt}), a crucial and time-intensive part of the workflow of documenting endangered languages. Retrieval augmented generation (RAG) incorporates an initial retrieval step, where LLMs query an external data source to gather relevant information before generating answers or text. This retrieval phase not only informs the subsequent generation process but also ensures that the responses are based on solid evidence, thereby improving the accuracy and relevance of the output. Figure~\ref{fig:explain-ex} shows an example of one sentence from the Mayan language Uspanteko, its true and predicted morpheme glosses, and the explanations produced by our RAG pipeline.
We test and compare the efficacy of two popular LLMs - Claude 3.5 Sonnet \cite{claude3.5} and GPT-4 \cite{achiam2023gpt}. In our experiments, Claude-3.5-Sonnet gives the most promising outputs, with both word and morpheme-level accuracies improving significantly over baselines for Uspanteko and Arapaho. 

LLMs also come with a known limitation: they are inherently data-hungry, relying on vast amounts of training data to achieve their impressive performance \cite{holmstrom-etal-2023-bridging}. This characteristic makes them less effective when dealing with small datasets, particularly prevalent in low-resource language contexts such as supporting documentation of endangered languages.
In these scenarios, leaner, tailor-made models 
seem to be preferred, offering better computational efficiency and flexibility.

In this paper, we specifically focus on the low-resource Uspanteko and Arapaho languages (Section~\ref{sec:usp}) as we have author-approved access to grammar resources and data for these languages. Our approach leverages the knowledge encapsulated in large models combined with these digitized grammars. In \citet{reid2024gemini}, Google's Gemini team demonstrated that they could fit an entire grammar for a low-resource language (Kalamang) in a single prompt owing to the massive context window size of Gemini 1.5.

While this zero-shot setting may work well for individual inputs, it is important to remember that the model must process the full grammar each time it receives a new prompt.

It is more computationally efficient and cost-effective to retrieve only the parts of the grammar  most relevant to the given query.

The RAG pipeline has been well-established for question-answering tasks, and this paper explores the capacity of LLMs to retrieve, interpret, and use only the relevant, retrieved parts of the grammar in a zero-shot setting to correct the output of a smaller model.

The process is not just about size reduction; it's a strategic transfer of linguistic capabilities, ensuring that the compact model inherits the teacher's strengths while remaining resource-efficient. In our baseline experimental setting, we simply call the LLM
at inference time to correct the output of the smaller token classification network. Experimenting further, we fine-tune the retrieval and re-ranking components in conjunction with the token classification model to boost performance. 

Furthermore, large language models can be prompted to explain their chain-of-thought \cite{wei2023chainofthought}, 
an approach with immense benefits for model explainability. 
Apart from correcting the output, we also generate a JSON object that contains descriptions of which
chunks were retrieved, how these chunks informed the final output, and the level of confidence the pipeline has in its final predictions. 
As seen in figure \ref{fig:explain-ex}, the LLM-generated explanations of the results and how RAG was used to achieve them are (largely) coherent and contextually relevant.

Our specific contributions include (1) development of a RAG pipeline to correct the predictions of a smaller model, (2) improving the usability of NLP models for language documentation by eliciting confidence scores and explanations for each prediction, (3) demonstration of significant performance improvements in low-resource language processing and a new SOTA for this task, and (4) a scalable approach that balances computational efficiency with linguistic accuracy and explainability.

\section{Background and Related Work}
\subsection{The glossing task} \label{sec:igt}
The specific task we address in this paper is morphological glossing, a component of the automatic production of interlinear glossed text (IGT).
IGT is a richly-annotated data format widely used in linguistics, especially as one product of work documenting and describing endangered languages. 

The data format, an example of which appears below,
consists of multiple interrelated tiers containing different types of linguistic information.
This Uspanteko example is representative of a common IGT configuration, with one tier for the original utterance, one for a morphological segmentation, one for a detailed labeling of the component morphemes, and one line with a translation into a language of wider communication.

\begin{small}
\ex 
\begingl
\gla xqil//
\glb x-$\emptyset$-q-il//
\glc COM-A3S-E1P-ver//
\glft `lo vimos' (`we saw it')//
\endgl
\xe
\end{small}

We focus specifically on the glossline, in which we see a mix of stem translations (e.g. \textit{ver} (in English, \textit{to see}) for the Uspanteko stem \textit{il}) and labels indicating morphosyntactic functions (e.g. \textsc{COM} indicates marking of completive aspect on the verb stem). Glossing is a sequence-to-sequence problem that can be approached in 2 ways. The first approach is to train a model to segment the data and then view it as a token classification task. The second is to view it as a translation problem with uneven input and output lengths, thus requiring an encoder-decoder model that can perform sequence-to-sequence conversion. In this paper, we use the former approach. The IGT task was the focus of a SIGMORPHON 2023 shared task. The task, resources, and previous models are described in detail in \citet{ginn-etal-2023-findings}. 

We directly use the segmented data in track 2 of the Sigmorphon 2023 glossing shared task to train our compact model. The corrective LLM only sees the segmented output of the compact model and the unsegmented text in the training data.\footnote{https://sigmorphon.github.io/sharedtasks/2023/}

\subsection{Related Work}

Analyses like those by \citet{conneau-etal-2020-unsupervised} reveal the inadequacies of large language models in capturing the nuances of less-common languages. These studies underline the necessity for specialized models that cater to the unique characteristics of low-resource languages. Furthermore, position papers like \citet{bender-koller-2020-climbing} critically analyze the bias and limitations in LLMs, advocating for more inclusive and adaptable language technologies. Another interesting approach to incorporating linguistic information is described in \citet{tziafas-etal-2023-improving}. They directly apply syntactic supervision at the pre-training stage to enhance the model with syntactic awareness. While this approach shows promising results, pre-training again requires large amounts of labeled data which is not available in a low resource setting.

Introduced by \citet{lewis2021retrievalaugmented}, Retrieval-Augmented Generation (RAG) represents a significant advancement in the field of large language models (LLMs) for enhancing generative tasks. By dynamically retrieving information from knowledge bases during inference, RAG effectively addresses issues such as the generation of factually incorrect content, often referred to as “hallucinations.” The integration of RAG into LLMs has been rapidly adopted, making it a crucial technology for enhancing chatbot capabilities and making LLMs more practical for real-world applications.

Naive RAG is the most basic form of retrieval augmented generation. The retrieve-read framework, which was described by \citet{ma2023query}, explains the process of indexing and vectorizing reference documents, retrieving relevant chunks based on vector similarity, and generating outputs based on compound prompts that combine the chunks and the query. The idea of RAG has since been expanded and adapted to several domains.
\citet{yan2024correctiveretrievalaugmentedgeneration} suggest a corrective RAG (CRAG) approach that incorporates a lightweight retrieval evaluator to test the quality and relevance of the retrieved content. The information is then filtered or accepted in the process of producing the final output. In our paper, we also add a corrective step in a different context. Instead of evaluating the retriever, we make the LLM itself generate confidence scores for each of its predicted outputs.

Other recent work on the IGT task takes various approaches. \citet{ginn2024glosslm} build a very large, multilingual corpus of IGT and use it to finetune a ByT5 model \cite{byT5}, achieving good results especially on languages not seen in training. 
\citet{he-etal-2024-wav2gloss} train models to extract IGT directly from audio data, and
\citet{ginn2024teachlanguagemodelsgloss} explore the use of in-context examples to teach LLMs to gloss low-resource language data. Using the same dataset we use, they find that LLM performance improves dramatically with targeted selection of examples. Even with no traditional training or fine-tuning, models like Gemini 1.5 Pro, Cohere's Command R+ and GPT-4o outperform transformer baselines. 
We do not explore few-shot prompting techniques, but it is possible that the performance of our corrective LLM can be further enhanced in this way.

\section{Methodology}\label{sec:methods}
Our approach combines the strengths of compact token classification models with the knowledge embedded in large language models (LLMs) and structured grammatical descriptions. The process involves several key steps:
\vspace{-0.3cm}
\begin{enumerate}
    \item \textbf{Initial glossing:} A compact token classification model (either RoBERTa or Bi-LSTM) generates an initial morphological gloss for the input sentence. 
    \vspace{-0.3cm}
    \item \textbf{Retrieval:} Relevant chunks of grammatical information are retrieved based on the input sentence and initial gloss.
    \vspace{-0.3cm}
    \item \textbf{Augmented generation:} An LLM uses the retrieved grammar chunks to correct and refine the initial gloss.
    \vspace{-0.3cm}
    \item \textbf{Explanation generation:} The LLM provides detailed explanations and confidence scores for each morpheme in the corrected gloss.
    \vspace{-0.3cm}
    \item \textbf{Modular optimization:} In an advanced version of our approach, we fine-tune the retrieval and token classification components together to optimize the entire pipeline.
\end{enumerate}
\vspace{-0.2cm}
We explore two main variants of this approach: a naive RAG method and a modular RAG method. 

\subsection{Using Naive RAG to correct the output of a smaller model}

The Naive Retrieval-Augmented Generation (RAG) approach enhances the performance of two compact token classification models trained for glossing of Uspanteko and Arapaho. The process begins by indexing and vectorizing reference grammar documents using a dense vector representation like BERT embeddings \cite{BERT}. Here, we use OpenAI embeddings \cite{achiam2023gpt}. 

Let $D = {d_1, d_2, \ldots d_n}$ be the set of $n$ grammar document chunks, where each $d_i$ is represented as a dense vector $v_i$ in a high-dimensional space $R^d$. We experimented with different chunk sizes and chunking strategies and found that a default chunk size of 400 characters with a 50-character overlap on either side provided the best results across 5 trials. We also tried chunking according to headings in the grammar, but this resulted in chunks of uneven sizes that did not optimally aid the retrieval process and contextual inference.

Given an input query $q$, in this case the sentence to be glossed along with the attempted prediction of the compact model, the retriever module computes the cosine similarity between the query embedding $v_q$ and each document chunk embedding $v_i$:
\[
\text{sim}(q, d_i) = \frac{v_q \cdot v_i}{\|v_q\| \|v_i\|}
\]
The top-$k$ most similar document chunks $\mathcal{D}_q = \{d_{q1}, d_{q2}, \ldots, d_{qk}\}$ are retrieved based on their cosine similarity scores. These chunks are concatenated with the original input query to form a compound prompt $P$:
\[
P = [q; d_{q1}; d_{q2}; \ldots; d_{qk}]
\]
The prompt $P$ is then fed into an LLM, which interprets the linguistic rules and morphological patterns described in the retrieved grammar excerpts $\mathcal{D}_q$. It uses this information to identify and correct potential errors in the glossing output $g_s$ generated by the smaller token classification model:
\[
g_c = \text{LLM}(P, g_s)
\]
where $g_c$ represents the corrected glossing sequence. To illustrate the correction process, let $f_s$ be the function learned by the smaller token classification model that maps the input sentence $x$ to the glossing output $g_s$.

The Naive RAG approach learns a corrector function $f_{\text{RAG}}$ that takes the original input $x$, the glossing output $g_s$, and the retrieved grammar chunks $\mathcal{D}_q$ to produce the corrected output $g_c$:
\[
g_c = f_{\text{RAG}}(x, g_s, \mathcal{D}_q)
\]
By leveraging the linguistic information retrieved from the grammar documents, the RAG model $f_{\text{RAG}}$ is able to refine the predictions of the base model $f_s$ and generate more accurate glossing sequences. The Naive RAG approach thus enables the smaller model to benefit from the vast knowledge captured by the LLM without the need for extensive fine-tuning or additional training data. This is particularly advantageous in low-resource scenarios where labeled data is scarce, as the LLM can provide valuable linguistic insights to guide the glossing process. [Refer to the appendix to see some of these generated explanations.]

\subsection{Generating Labeling Justifications and Confidence Scores}

In addition to correcting glossing labels, our RAG pipeline also generates explanations justifying the corrections made. We achieve this by prompting the LLM to provide a chain-of-thought reasoning trace that justifies the decision-making process behind the corrections. The LLM is prompted with an instruction $I$ that requests a justification $J$, an explanation of how RAG was used $R$, and a confidence score $C$ for corrected glossing output $g_c$:
\[
[J, R, C] = \text{LLM}(I, P, g_s, g_c)
\]
The justification $J$ is a natural language explanation that describes the grammar rules and morphological patterns retrieved from the grammar excerpts $\mathcal{D}_q$ and how they informed the corrections made to the glossing sequence. This explanation can be modeled as a set of reasoning steps:
\[
J = [r_1, r_2, \ldots, r_m]
\]
where each $r_i$ represents a single reason that links the retrieved linguistic information to the specific corrections made in $g_c$. This set of $r_i$ is not necessarily sequential.

To quantify the model's confidence in the corrected output, a confidence score $C$ is generated. This score reflects the LLM's certainty in the accuracy of the final glossing sequence based on the retrieved grammar rules and the original output $g_s$. When asked how it assigned confidence scores, this was Claude's response:

\textit{"Confidence scores were assigned based on how closely each word or morpheme matched information provided in the grammar document. Higher scores (closer to 1.0) indicate a strong direct match, while lower scores (closer to 0.5) indicate a more tentative match based on context or inference."}

Through these confidence scores and justifications, the RAG pipeline boosts the interpretability of the model's predictions. The explanations offer insights into the linguistic reasoning behind the corrections, allowing users to understand why certain changes were made. This can be crucially important for documentary linguists who may be reluctant to use NLP tools due to concerns about reliability and a lack of interpretibility.

\subsection{Modular RAG: Training the retriever with the sequence to sequence model}

Due to the size of our grammars, it makes sense to further train our retriever and rank the retrieved content based on its relevance to the query. This is an extension of the previously described Naive RAG approach. 
The modular RAG approach allows for a more sample-efficient utilization of the available grammar resources. By learning to retrieve and prioritize the most relevant excerpts, the model can focus on the linguistic information that is most beneficial for each specific input, rather than processing the entire grammar at once.

\begin{enumerate}
    \item \textbf{Initial Retrieval:} The process begins by retrieving $k$ context chunks from the grammar based on relevance to the input query.
    
    \item \textbf{Retrieval Module Training:} The retrieval module, which is a different instance of the RoBERTa model, is fine-tuned based on the performance of the LLM outputs. The input to the retrieval module consists of all the initially retrieved chunks of context, and the output is a relevance vector $r = [r_1, r_2, ..., r_k]$, where $r_i \in [0, 1]$ indicates the relevance score for the $i^{th}$ chunk of context.
    
    \item \textbf{Final Context Selection:} Out of the $k$ retrieved pieces of context, only the top-$n$ most relevant pieces are selected for use in the final LLM prompt.
\end{enumerate}

Let $f_s$ be the token classification model that maps input sentence $x$ to the initial glossing output $g_s$. The retriever module $f_r$ is trained to select the top-$n$ relevant grammar chunks $D_q$ based on the input sentence $x$ and the initial glossing output $g_s$:

\[
D_q = f_r(x, g_s, k, n)
\]

where $k$ is the initial number of retrieved chunks and $n$ is the final number of selected chunks $(n \leq k)$.

The selected grammar chunks $D_q$ are concatenated with the input sentence $x$ and the initial glossing output $g_s$ to form a prompt $P$:

\[
P = [x; g_s; D_q]
\]

The prompt $P$ is then fed into the LLM to generate the corrected glossing sequence $g_c$:

\[
g_c = \text{LLM}(P)
\]

During training, the retriever $f_r$ and token classification model $f_s$ are jointly optimized using a combined loss function:

\[
L = L_s(g_c, g_t) + \alpha \cdot L_r(D_q, D_t)
\]

where:
\begin{itemize}
    \item $L_s$ is the sequence loss between the corrected glossing sequence $g_c$ and the ground truth glossing labels $g_t$.
    \item $L_r$ is the retrieval loss that encourages the retriever to select relevant grammar chunks. It is implemented as a ranking loss between the retrieved chunks $D_q$ and the chunks that led to the best LLM performance $D_t$.
    \item $\alpha$ is a hyperparameter that controls the weight of the retrieval loss.
\end{itemize}

By jointly optimizing the retriever and token classification components, the modular RAG approach enables the model to learn to identify the most relevant grammar information and effectively incorporate it into the glossing process. The retriever learns to select excerpts that are most pertinent to the input sentence and initial gloss output.

During inference, the trained retriever $f_r$ is used to select the top-$n$ relevant grammar chunks $D_q$ for each input sentence $x$ and initial glossing output $g_s$. These selected chunks are provided to the LLM to generate the corrected glossing sequence $g_c$.

\subsection{Baseline Glossing Models}\label{sec:baselines}

Our expectation is that the best results for this task will be achieved by combining a model that exploits all available training data (the compact transformer or LSTM model) with the analytical power of modern LLMs. 

More specifically, we use two baseline glossing models, both of which have been shown to achieve strong performance in the standard setting for the glossing task \cite{ginn-etal-2023-findings}. Following this setting, we model the production of IGT as a token classification task rather than as a sequence-to-sequence task.
Specifically, we experiment with two baseline models: one using RoBERTa \cite{liu2019roberta}, the second using a Bi-LSTM architecture. 
By exploiting the contextual information captured by these architectures, we aim to obtain accurate predictions of the morphological labels for each token in the input sentences. These predicted labels are then used as the initial glossing output in the RAG framework.

For the RoBERTa baseline, we use the same setting as the baseline for the IGT shared task, as described in \citet{ginn-etal-2023-findings}.
The input sentences are tokenized and encoded using the RoBERTa tokenizer and encoder. The encoded representations are then passed through a linear classification layer to predict the morphological labels for each token. 

For the Bi-LSTM model, input sentences are first tokenized and converted into word embeddings. These embeddings are then fed into the Bi-LSTM layer to obtain the contextualized token representations. A linear classification layer is applied on top of the Bi-LSTM outputs to predict the morphological labels for each token. 
Both models are trained using cross-entropy loss and optimized using the Adam optimizer. We use an adaptive learning rate and early stopping to ensure a better fit to the data.

We additionally compare with two different LLMs used in a single-model, zero-shot RAG architecture. In this setting, the LLMs are solely responsible for the glossing output, rather than correcting the output of a predecessor model.

\section{Uspanteko and Arapaho: Data and Grammars}\label{sec:usp}

\begin{table*}[th!]
    \small
    \centering
    \renewcommand{\arraystretch}{1.5} 
    \begin{tabular}{|l|c|c||c|c|} \hline
             & \multicolumn{2}{c|}{\textbf{Uspanteko}} & \multicolumn{2}{c|}{\textbf{Arapaho}} \\ \cline{2-5}
         \textbf{Model} & Word-level & Morpheme-level & Word-level & Morpheme-level \\
             & Accuracy & Accuracy & Accuracy & Accuracy \\ \hline
        GPT-4 Baseline (Zero-shot RAG) & 42.21 & 51.88 & 48.47 & 53.48 \\ \hline
        Claude Baseline (Zero-shot RAG) & 38.40 & 42.21 & 49.91 & 58.60 \\ \hline 
        Previous SOTA (Shared task) & 78.46 & 84.51 & 85.87 & 91.37 \\ \hline \hline 
        RoBERTa Baseline & 76.55 & 82.48 & 85.44 & 91.11 \\ \hline
        RoBERTa + Claude (Train + RAG) & \textbf{79.21} & \textbf{84.84} & \underline{\textbf{86.82}} &\underline{\textbf{93.74}} \\ \hline
        RoBERTa + GPT-4 (Train + RAG) & 78.41 & 81.49 & 85.51 & 91.43 \\ \hline
        RoBERTa + Claude (Modular RAG) & \underline{\textbf{81.12}} & \underline{\textbf{85.02}} & \textbf{83.98} & \textbf{90.26} \\ \hline
        RoBERTa + GPT-4 (Modular RAG) & 79.44 & 82.98 & 82.41 & 88.68 \\ \hline \hline
        Bi-LSTM Baseline & 71.28 & 73.90 & 76.41 & 80.44 \\ \hline        
        Bi-LSTM + Claude (Train + RAG) & 77.47 & 80.21 & 79.12 & 85.44 \\ \hline
        Bi-LSTM + GPT-4 (Train + RAG) & 73.17 & 78.23 & 74.16 & 81.31 \\ \hline 
        Bi-LSTM + Claude (Modular RAG) & 78.26 & 82.22 & 81.24 & 85.89 \\ \hline
        Bi-LSTM + GPT-4 (Modular RAG) & 74.12 & 78.99 & 76.77 & 82.18 \\ \hline
    \end{tabular}
    \caption{Comparison of all model performances for Uspanteko and Arapaho. Averaged over 5 runs. Highest scores for each model type (naive, modular) are in boldface; overall high scores underlined. Naive RAG retrieves up to 6 relevant chunks of context while Modular RAG restricts this to the top 3 chunks.}
    \label{tab:results}
\end{table*}

Uspanteko is an endangered Mayan language spoken primarily in Guatemala. 
It is an ergative-absolutive language with moderately complex concatenative morphology.
Much morphological inflection occurs on the verb stem, which takes both prefixes and suffixes and inflects for person, number, participant role, tense/aspect/mood, and voice, with a final status suffix. Arapaho is an endangered Algonquian language spoken by several communities in the Western United States. The language has free word order, polysynthetic and agglutinating morphology, and especially complex verbal morphology \cite{cowell2011arapaho}.

\paragraph{Data.} We use the Uspanteko and Arapaho IGT datasets provided as part of the 2023 SIGMORPHON shared task \cite{ginn-etal-2023-findings}, licensed under CC BY-NC 4.0, and we use the data in accordance with the uses intended as part of the shared task. The Uspanteko dataset has about 11,000 usable sentences and about 80 unique morphological function labels. 

The average sentence is 4.37 words, with many multi-morphemic words.
The Arapaho dataset is much larger, consisting of 39,500 sentences (5.4 words on average per sentence) in the training set and 5000 in the dev set.

For Uspanteko, we use a very short (10 page) grammatical description, in Spanish, from the beginning of an  Uspanteko-Spanish dictionary \cite{uspanteko_grammar}. For Arapaho, we use a  500-page reference grammar authored by Andrew Cowell and Alonzo Moss, Sr. \cite{cowell2011arapaho}.

\section{Experiments and results}

Table~\ref{tab:results} shows results for all experimental settings, as well as the previous state-of-the-art for each language, as reported in \citet{ginn-etal-2023-findings}. 
The two LLM-only baselines perform well below the glossing baselines (RoBERTa and Bi-LSTM, see~\ref{sec:baselines}) and all other models.
For each of the two glossing baselines, we compare our naive and modular RAG models (see~\ref{sec:methods}), separately in combination with Claude and GPT-4.
We aim to evaluate which LLM is most effective at correcting the glossing output of the smaller token classification network, given retrieved grammar excerpts. Before evaluation, we perform post-processing to correct some common punctuation errors in the LLM output.

We evaluate on both word-level and morpheme-level accuracy metrics as described in \cite{ginn-etal-2023-findings}. These metrics are computed by comparing the corrected glossing sequences $g_c^{LLM}$ with the ground truth glossing labels $g_t$ for each input sentence $x$ in the test set. We manage to beat the previous SOTA with modular RAG for Uspanteko and naive RAG for Arapaho.

We see that a RAG approach combining the RoBERTa baseline with Claude consistently performs best. The Bi-LSTM model performs reasonably well in most cases, although it consistently trails RoBERTa. Selective retrieval seems to help more with Uspanteko than Arapaho. In fact, we see a performance drop when we train the retriever with Arapaho. Modular RAG retrieves a smaller, more focused set of grammar chunks than naive RAG. It is possible that this reduced set fails to capture all the information needed to inform the Arapaho gloss correction process, resulting in a small accuracy drop.

\section{Qualitative analysis: usability}\label{sec:qual}

\begin{table*}[]
\small
\begin{tabular}{|l|l|l|c|}\hline
\textbf{Type} & \textbf{Explanation}                                     & \textbf{Example} & \textbf{Frequency} \\ \hline
content       & true mismatches between expected output and model output & FUT for PAST     &      30              \\ \hline
form          & variation in form only; likely resolvable by users       & EXIST for EXS    &     18               \\ \hline
specificity & generated output is more or less specific than expected output               & NOM for SAB (abstract noun) & 2 \\ \hline
category    & generated tag where lexical output is expected, or vice versa                & PROHIB for `eat something'  & 39 \\ \hline
presence    & generated output contains spurious labels,  & PAST-NEG for PAST           &  10 \\
 & or has fewer labels than expected & & \\ \hline
 unk & model generates ?, or replaces ? with a guess & SREL for ? & 4 \\ \hline
\end{tabular}
\caption{Error types and frequencies across 50 randomly-selected instances.}
\label{tab:errors}
\end{table*}
\begin{table*}[t]
\small
\centering
\begin{tabular}{|l|c|c|c|c|c|c|} \hline
 & 
 \textbf{pre-LLM errors} &
  \textbf{corr/inc/part} &
  \textbf{new errors} &
  \textbf{\% corr. expl.} &
  \textbf{exp. quality (1-5)} &
  \textbf{ret. quality (1-5)}\\ \hline
\textbf{Arapaho} &
 21  &
  7 / 10 / 4  &
 7  &
 82.55\%  &
  3 &
  1.98 \\ \hline
\textbf{Uspanteko} &
 23  &
 16 /  7 / 0  &
  9 &
  81.38\% &
  3.19 &
 2.42 \\ \hline  
\end{tabular}
\caption{Manual analysis of model output, 14 Arapaho/16 Uspanteko examples. We count model corrections that are \textbf{correct}, \textbf{incorrect}, and \textbf{partially correct}, as well as \textbf{new errors} introduced by the corrective LLM. We also rate the average \textbf{explanation correctness} and quality of both \textbf{explanations} and \textbf{retrieved chunks}. Details in Appendix C.}
\label{tab:qual}
\end{table*}

This system is designed to support linguists and others performing the work of interlinear glossing. The explanations generated by the model improve interpretability, as they provide an opportunity for human users to get some insight into the model's decision-making process. 
The best evaluation of the usability of our system would come from proper user studies, which we have begun and will report on in later work. For the present paper, we perform two manual analyses, both performed using outputs from our Modular RAG pipeline with RoBERTa + Claude.

\paragraph{Glossing error types.}\label{sec:qualError}
Initial inspection of system outputs showed that, in some cases, the LLM proposes a corrected gloss that is close to the expected output without being identical, resulting in a ding to automatically-evaluated performance. For example, the model sometimes outputs \textsc{3S} when the expected tag is \textsc{3.S}. We randomly select 50 instances across the two datasets and evaluate them for error types. Table~\ref{tab:errors} explains our error types and their frequency across this sample; we identify 103 errors across the 50 instances. Arapaho sentences have an average of 2.2 errors per sentence, with 1.9 for Uspanteko. 

Category-type errors, where the model generates a tag instead of a lexical item, or vice versa, are most common, followed by content-type errors, which we consider ``true'' glossing errors. The error types form and specificity are those which we expect to be easily interpreted and corrected by human users; these account for roughly 19\% of the errors. See Appendix~\ref{app:glossErr} for error subtypes.

\paragraph{Quality of explanations.}\label{sec:qualExpl}
Our second manual analysis concerns the quality and relevance of the explanations provided by the LLM.

Examples in App.~\ref{app:expl} show the two-part structure of the explanations: 1) explanation of the presumed meaning of the morphemes, 2) explanation of which parts of the grammar were retrieved and used to make glossing decisions.

We randomly select 30 instances. For each, we collect the original text, expected gloss, output of the initial glossing model, LLM-corrected output, and the complete set of explanations and retrieved grammar chunks from the RAG pipeline. A professional linguist then analyzes the number of pre-LLM errors, how many are addressed correctly/incorrectly/partially correctly, the percentage of correct morpheme explanations, the subjective quality of the RAG explanations, and the subjective quality of the retrieved grammar chunks, the latter two on a Likert scale (1-5). (Details in App.~\ref{app:qualExpl}.)

The results appear in Table~\ref{tab:qual}. On average, the Uspanteko corrections are more accurate, with a similar number of new errors being introduced for both languages. Model explanations for individual morphemes are largely correct, and the chunks retrieved for Uspanteko are slightly higher quality. We note that there is a clear difference in the nature of the two grammars. The Arapaho grammar is a full and complex reference grammar, and the Uspanteko grammar is a sketch, using simpler explanations in a more compact presentation. This initial analysis suggests the need for a deeper exploration into linguistic reference materials of different types and their use in RAG.
Arapaho morphology is also significantly more complex than Uspanteko morphology, increasing the complexity of the  task.

\section{Conclusion}
This study demonstrates the effectiveness of a Retrieval-Augmented Generation (RAG) framework in enhancing the performance of compact models for morphological glossing in low-resource language contexts. By leveraging the interpretive power of Large Language Models (LLMs) and the structured knowledge contained in grammatical descriptions, we achieve a new state-of-the-art for both languages investigated. A second advantage is the interpretability provided by LLM-generated explanations, which is crucial for building trust in the system's outputs and facilitating the use of NLP tools in language documentation efforts. 

The RAG approach (combining a RoBERTa baseline with Claude) consistently outperforms other configurations, achieving the highest word- and morpheme-level accuracies for both languages. This framework effectively bridges the gap between the limited training data available for low-resource languages and the rich linguistic knowledge encoded in grammatical descriptions. The ability of LLMs to provide detailed explanations and confidence scores for each morpheme adds a layer of interpretability to the glossing process, potentially increasing the utility of these tools for documentary linguists.
Even with minimal grammatical resources, as for Uspanteko, the RAG approach shows notable improvements over baseline models.

These findings suggest that the integration of linguistic knowledge through RAG can be a powerful approach for improving NLP tasks in low-resource settings. By combining the strengths of compact, trainable models with the vast knowledge encoded in LLMs and structured grammatical descriptions, we can create more accurate and interpretable tools for language documentation and analysis.

For future work, we would like to investigate additional languages, explore more sophisticated retrieval mechanisms, incorporate additional linguistic resources (as in \citet{zhang2024hire}), and optimize our LLM selection and fine-tuning approaches. Near term, we plan to implement the same framework using an open-source LLM.

\section{Limitations}

While the proposed RAG framework for morphological glossing demonstrates promising results, there are several limitations to consider:

\begin{enumerate}
    \item \textbf{Dependency on grammar quality:} The effectiveness of the RAG pipeline heavily relies on the quality and comprehensiveness of the available grammar documents. If the grammar descriptions are incomplete, inconsistent, or contain errors, the retrieved excerpts may not provide accurate or sufficient information to guide the glossing corrections. This can lead to sub-optimal performance of the RAG model.
    \item \textbf{Limited expressiveness of grammars: }The linguistic rules and patterns described in grammar documents may not capture all the nuances and exceptions present in the target language. Some morphological phenomena may be too complex or irregular to be fully expressed in a concise set of rules. This limitation can hinder the RAG model's ability to generate accurate glossing labels for such cases. This is especially true in the case of our relatively small Uspanteko grammar.
    \item \textbf{Scalability to larger datasets:} The current experiments focus on low-resource languages with relatively small datasets. While the RAG approach is designed to be data-efficient, its performance and computational requirements when applied to larger datasets or more diverse language families remain to be investigated. The retrieval and processing of grammar excerpts may become more challenging as the size and complexity of the data increases.
    \item \textbf{Generalization to unseen languages:} The RAG pipeline has been evaluated on specific low-resource languages, such as Uspanteko and Arapaho. However, its generalization capability to other unseen languages with different morphological typologies is not extensively tested. The effectiveness of the approach may vary depending on the similarity of the target language to the languages used in training and the availability of suitable grammar resources.
    \item \textbf{Reliance on proprietary models:} We currently use two proprietary LLMs for these experiments. Once we have the appropriate compute infrastructure established, we plan to implement the same architecture using an open-source model.
    \item \textbf{Single evaluator for manual qualitative analysis:} So far we have performed only a small-scale, somewhat subjective analysis of the quality of the generated explanations and the relevance of retrieved grammar chunks for the sentences being glossed. We are currently building a web interface to allow linguists and other users to interact and evaluate model outputs. Once finished, we will expand this analysis to a more robust user study with multiple users and multiple languages.
\end{enumerate}

\section{Ethical Considerations}

Although generative AI is clearly helpful in making the process of language documentation more explainable and efficient, it can also be deceptively convincing in its justifications. Through the initial prompt, the LLM is primed to report any uncertainties and assign confidence scores, but as seen in the qualitative analysis, this is in no way a foolproof approach. The RAG-driven system is not ready to be used by language learners or for large-scale language documentation tasks without further testing.

We also recognize that we use sensitive language data and resources to obtain our results. All our data is part of the publicly available Sigmorphon 2023 shared task data as mentioned in section \ref{sec:usp}, and we have permission from the producers of the two grammars to use them for research purposes. We will continue to seek express permission from communities and authors before using low-resource data and grammars for future experiments.

\bibliography{custom}

\appendix

\section{Sample explanations}
\label{app:expl}

\subsection{Uspanteko Example}
Given the Uspanteko sentence "t wi' tene' qleen tob'ool laq b'ee", Claude produced the following corrected gloss: "EXST talvez NEG-existir ayudar-NMLZ PREP camino".
In addition to the corrected gloss, Claude generated a detailed explanation for each morpheme, along with confidence scores:

\begin{itemize}
    \item \textit{'wi" is an existential predicate (EXIST) indicating the presence or existence of something. (confidence: 0.9)}
    \item \textit{'tene" is a particle meaning 'maybe' or 'perhaps'. (confidence: 0.7)}
    \item \textit{'qleen' is a negative existential predicate (NEG-exist), indicating the non-existence of something. (confidence: 0.9)}
    \item \textit{'tob'ool' can be broken down as: 'tob" (verb root meaning 'to help'), '-ool' (suffix nominalizing the verb, NMLZ), yielding a noun meaning 'help' or 'assistance'. (confidence: 0.8)}
    \item \textit{'laq' is a preposition meaning 'at' or 'in'. (confidence: 0.8)}
    \item \textit{'b'ee' is a noun meaning 'path', 'road', or 'way'. (confidence: 0.9)}

\end{itemize}

Claude also provided information on how the RAG approach was used to arrive at the corrections:

\begin{itemize}
    \item \textit{The existential predicate 'wi" and the negative existential predicate 'qleen', both of which appeared in the example sentences.}
    \item \textit{The information about the suffix '-ool' nominalizing verbs (NMLZ), which was mentioned in the 'Estructura de las entradas' (Structure of the entries) section.}
    \item \textit{The noun 'b'ee' meaning 'path', which appeared in one of the example sentences.}
    \item \textit{The meanings of 'tene" and 'laq' were inferred from context, as they are not explicitly defined in the grammar document.}
\end{itemize}

\subsection{Arapaho Example}
Given the Arapaho sentence "Hookoh neniih'ei3i'okuno' hoo3oo'o' hiniito'eino notoone hiihoowuno'useenino ", Claude produced the following corrected gloss: "because IC.scattered-sit-1PL what 3S-relative-OBV.PL 1S-people NEG-arrive-3PL".
In addition to the corrected gloss, Claude generated a detailed explanation for each morpheme, along with confidence scores:

\begin{itemize}

    \item \textit{\textbf{hookoh}: glossed as \textit{because}, seems to be a conjunction (confidence: 0.9)}
    
    \item \textit{\textbf{neniih'ei3i'okuno'}: a complex verb form (confidence: 0.8)}
    \begin{itemize}
        \item \textit{\textbf{neniih'ei-}: possibly an initial change (IC) form of \textit{scattered}
        \item \textbf{-3i'oku-}: likely means \textit{sit}
        \item \textbf{-no'}: 1PL (first person plural) marker}
    \end{itemize}
    
    \item \textit{\textbf{hoo3oo'o}: glossed as \textit{what}, likely an interrogative pronoun (confidence: 0.9)}
    
    \item \textit{\textbf{hiniito'eino}: appears to be a noun phrase (confidence: 0.7)}
    \begin{itemize}
        \item \textit{\textbf{hini-}: possibly 3S (third person singular) prefix
        \item \textbf{-ito'ei-}: root meaning \textit{relative}
        \item \textbf{-no}: OBV.PL (obviative plural) marker}
    \end{itemize}
    
    \item \textit{\textbf{notoone}: glossed as \textit{1S-people}, likely a possessed noun (confidence: 0.8)}
    \begin{itemize}
        \item \textit{\textbf{no-}: 1S (first person singular) possessive prefix
        \item \textbf{-toone}: root meaning people}
    \end{itemize}
    
    \item \textit{\textbf{hiihoowuno'useenino}: a complex verb form (confidence: 0.8)}
    \begin{itemize}
        \item \textit{\textbf{hiihoowu-}: NEG (negation) prefix
        \item \textbf{-no'usee-}: verb root meaning arrive
        \item \textbf{-nino}: 3PL (third person plural) marker}
    \end{itemize}
\end{itemize}

Claude also provided information on how the RAG approach was used to arrive at the corrections:

\begin{itemize}
    \item \textit{The basic morphosyntax structure provided insight into the order of elements in the verb complex.}
    
    \item \textit{The document mentions person markers and their positions, which helped in identifying \textbf{-no'} as 1PL and \textbf{-nino} as 3PL.}
    
    \item \textit{The negation prefix \textbf{ihoowu-} is mentioned, which appears here as \textbf{hiihoowu-}.}
    
    \item \textit{The document discusses obviative plural markers, supporting the interpretation of \textbf{-no} in \textit{hiniito'eino}.}
    
    \item \textit{The contraction rules mentioned weren't directly applicable here but provided context for understanding morpheme boundaries.}

\end{itemize}
\section{Details of qualitative analyses}
\label{app:qual}

\subsection{Glossing error types}\label{app:glossErr}

In section~\ref{sec:qualError} we present an analysis of the types of glossing errors made by our best model, across 50 randomly-selected sentences.
The figures presented there are for high-level error categories.
During the analysis, we consider a number of subtypes for each high-level error type. 

Table~\ref{tab:subtypes} (next page) shows the complete set of error types and subtypes, with frequencies, examples, and descriptions.

\begin{table}[]
\begin{tabular}{|l|l|}
\hline
  & \textbf{usefulness/correctness}                  \\ \hline
1 & all explanations incorrect and/or unuseful       \\ \hline
2 & most explanations incorrect and/or unuseful      \\ \hline
3 & about half of explanations correct and/or useful \\ \hline
4 & most explanations correct and/or useful          \\ \hline
5 & all explanations correct and/or useful           \\ \hline
\end{tabular}
\caption{Likert scale used to score morpheme explanations provided by the corrective LLM.}
\label{tab:likertExpl}
\end{table}
\begin{table}[]
\begin{tabular}{|l|l|}
\hline
  & \textbf{quality/relevance}                      \\ \hline
1 & all explanations unhelpful or misleading        \\ \hline
2 & most explanations unhelpful or irrelevant       \\ \hline
3 & about half of explanations relevant and helpful \\ \hline
4 & most explanations relevant and helpful          \\ \hline
5 & all explanations relevant and helpful           \\ \hline
\end{tabular}
\caption{Likert scale used to score RAG explanations provided by the corrective LLM.}
\label{tab:likertRetr}
\end{table}

\begin{table*}[]
\small
\begin{tabular}{|l|l|l|l|c|} \hline 
\textbf{type}              & \textbf{subtype} & \textbf{example}    & \textbf{notes}                                                       & \textbf{frequency} \\ \hline \hline
\textbf{content}     & wholeDiff & FUT for NEG            & single tag wrong, output is                                   &  13 \\
 & & & entirely different linguistic dimension     & \\ \hline
\textbf{}         & wholeSame        & FUT for PAST        & single tag wrong, output is               &                  8  \\ 
 & & & same linguistic dimension  & \\ \hline
\textbf{}         & partial          & E3S for E3P         & one part of compound tag is incorrect                                & 8                   \\ \hline
\textbf{}         & multiple         & 0S for 3PL          & all parts of compound tag are incorrect                              & 1                   \\ \hline\hline
\textbf{form}        & variant   & EXIST for EXS          & output has generated a plausible variant                    &  5 \\ 
 & & & not in the tagset (could be in   the grammar) & \\ \hline
\textbf{}            & similar   & IMP for IMPER          & output is incorrect tag, similar to correct                                  &  2 \\ 
& & & tag, both are in the tagset & \\ \hline
\textbf{}            & punct     & 3.S for 3S, 3-S for 3S & only difference is punctuation (could be              &  9 \\ 
& & & missing, could be spurious,   could be replacement) & \\ \hline
\textbf{}         & case             & PAUSE for pause     & difference is case (which is  &                  2  \\ 
& & & potentially meaningful in this setting) & \\ \hline \hline
\textbf{presence} & extra            & PAST-NEG for PAST   & output contains spuriously generated material                        & 8                   \\ \hline
\textbf{}         & missing          &                     & output is missing a tag                                              &  2                  \\ \hline \hline
\textbf{specificity} & hyper     & NOM for SAB            & generated output is less specific than expected                   &  1 \\ 
& & & tag (e.g. nominal for   abstract noun)  & \\ \hline
\textbf{}            & hypo      & DET for PART           & generated output is more specific than expected  & 1 \\ 
& & & tag (e.g. DET could be   one of many types of particles) & \\ \hline \hline
\textbf{category} & 2lex             & so.that for DETACH  & generated output has lexical translation instead of tag              &    17                \\ \hline
\textbf{}         & 2tag             & PROHIB for eat.s.t. & generated output has tag instead of lexical translation              &      22              \\ \hline \hline
\textbf{unk}      & unk              & ? for SC            & model generates ?                                                    &        2            \\ \hline
\textbf{}         & guess            & SC for ?            & model guesses where original gloss has question marks                &         2          \\ \hline
\end{tabular}
\caption{Glossing error analysis types and subtypes, together with frequencies across 50 sentences.}
\label{tab:subtypes}
\end{table*}

\subsection{Quality of explanations}\label{app:qualExpl}

Our process for analyzing the quality of explanations provided consisted of five steps.

\begin{enumerate}
\item Compare glossing output of the baseline token classification model to the expected (gold standard) glossing output, counting the number of errors at the morpheme level.
\item Compare the LLM-corrected to the baseline output. For each error in the baseline output, determine whether the LLM made a correct correction, an incorrect correction, or a partially correct correction. In addition, look for new errors introduced by the corrective LLM.
\item For the set of morpheme explanations, mark each as correct, partially correct, or incorrect. Determine the percentage of correct explanations by comparing to the expected gloss, with partially correct explanations receiving 0.5 points.
\item Rate the set of explanations about how RAG was used according to their usefulness and/or correctness, using the scale in Table~\ref{tab:likertExpl}.
\item For each retrieved grammar chunk, rate its quality/relevance for the example being glossed, using the scale in Table~\ref{tab:likertRetr}. Compute the average score across all retrieved grammar chunks.
\end{enumerate}

\end{document}